\documentclass[letterpaper, 10 pt, conference]{ieeeconf}  
\IEEEoverridecommandlockouts
\usepackage[T1]{fontenc}
\usepackage[latin1]{inputenc}
\usepackage{babel}
\usepackage[font=small,labelfont=bf,tableposition=top]{caption}
\usepackage{blindtext}
\usepackage{cuted}
\usepackage{graphicx}
\usepackage{subfiles}
\usepackage{cite}
\usepackage{amsmath}
\usepackage{algorithm}
\usepackage{algpseudocode}
\usepackage{subcaption}
\usepackage[export]{adjustbox}
\usepackage[colorlinks,citecolor=blue,urlcolor=blue,bookmarks=true,hypertexnames=true]{hyperref} 

\title{Contact simulation of a 2D Bipedal Robot kicking a ball}
\author{Alphonsus Adu-Bredu$^{1}$ 
\thanks{$^{1}$Alphonsus Adu-Bredu is with the Robotics Institute, University of Michigan, Ann Arbor, MI, USA.
        {\tt\small adubredu@umich.edu}}
}

\begin{document}
\maketitle
\noindent

\begin{strip}
\centering\noindent 
\includegraphics[width=1.0\linewidth]{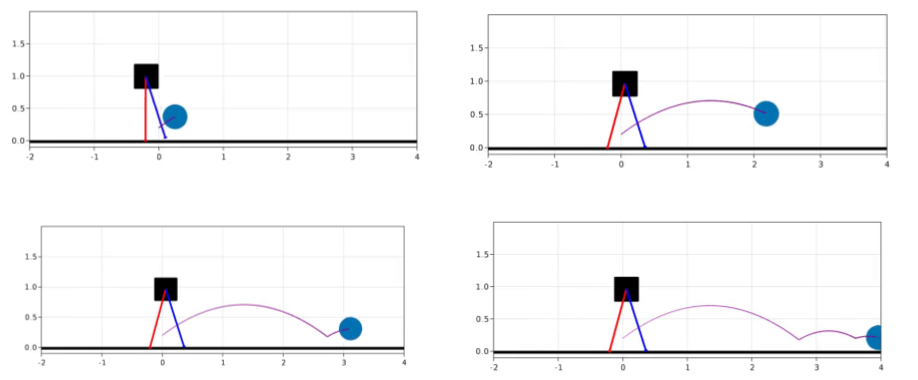}
\captionof{figure}{Biped robot walks to towards the ball and kicks it.}
\end{strip}

\begin{abstract}
    This report describes an approach for simulating multi-body contacts of actively-controlled systems. In this work, we focus
    on the controls and contact simulation of a 2-dimensional bipedal robot kicking a circular ball. The Julia code and more video results of
    the contact simulation can be found at this URL: \href{https://github.com/adubredu/contact_sim.jl}{https://github.com/adubredu/contact\_sim.jl}
\end{abstract}
\section{Introduction}
\subsection{Motivation and Context}
Contact resolution is useful for actively-controlled rigid-body systems that make and break contact with their environment. The ability to resolve contact forces after collision with a rigid body enables the prediction of the system's motion after impact. The resolved contact forces can then be accounted for in the system's feedback controllers in order to keep the system stable and functional. 

\subsection{Problem Statement}
In this project, we simulate multi-body contacts of an actively-controlled rigid-body system. More specifically, we focus on the contact simulation and control of a 2-dimensional bipedal robot kicking a circular ball. We assume plastic contacts for the foot of the 2D bipedal robot as it walks on the flat ground and elastic contacts for the circular ball as it bounces on the ground after being kicked.

\subsection{Overview}
The remaining sections are organized as follows; In Section \ref{subsec:lcp}, we describe our general approach for contact resolution for rigid bodies. Next, in Sections \ref{subsec:ball} and \ref{subsec:bipedcontact} , we describe how we apply contact resolution for simulating contacts of a circular ball and a 2D bipedal robot respectively. Finally, we describe how we incorporate resolved contact forces in controlling the locomotion of the 2D bipedal robot. We provide qualitative and quantitative results in Section \ref{sec:experiments} and conclude the report in Section \ref{sec:conclusion} 
\section{Literature Review}
Contact resolution has been studied extensively both in the field of robot manipulation \cite{hogan2020feedback, wang2021gelsight, suomalainen2021survey} and robot locomotion \cite{bouyarmane2012humanoid, carpentier2017learning, liljeback2009modelling}. This work aims to simulate contacts of an actively-controlled bipedal robot that interacts with a passive rigid body. We resolve contacts using the linear complementarity formulation as has been pursued in works like \cite{todorov2010implicit, erleben2013numerical}.
\section{Methodology}\label{sec:methodology}

\subsection{Linear Complementarity Formulation for Contact Resolution}\label{subsec:lcp}
Since most contact mechanics constraints have complementarity relationships, one common way of resolving contact forces after impact is to formulate contact resolution as a Linear Complementarity Problem (LCP). Linear Complementarity Problems have the form \begin{equation}
\begin{aligned}
\max_{z} \quad & z^T \cdot (V\cdot z + p)\\
\textrm{s.t.} \quad & V\cdot z + p \geq 0\\ 
 & z \geq 0   \\ 
\end{aligned}
\end{equation}

where $V \cdot z + p \geq 0$ and $z \geq 0$ have the complementarity relationship $V \cdot z + p \geq 0 ~\bot~ z \geq 0$, which implies that, element-wise, at least one of the expressions on the other side of $\bot$ takes on a value of $0$.

To formulate contact mechanics constraints of a rigid body as an LCP, we first discretize the dynamics equation 
\begin{equation}
\begin{aligned}
    M(q) \dot{v} + c(v,q)+g(q) = f_e\\
    v = \dot{q}
\end{aligned}
\end{equation}

by taking first-order zero-hold discrete-time approximation. We then rearrange the terms to express the equations in terms of the position and velocity of the body in the next time step as
\begin{equation}
\begin{aligned}
    v_{t+1} = v_t + \Delta t M^{-1}(q_t)k(v_t, q_t) + f_{e, t} \\
    q_{t+1} = q_t + \Delta tv_t
\end{aligned}
\end{equation}
Finally, we introduce the reaction forces due to contact and their relevant constraints to form the system of equations and constraints
\begin{equation}
\begin{aligned}
    v_{t+1} = v_t + \Delta t M^{-1}(q_t)(k(v_t, q_t) + f_{e, t}  + J_c(q_t)f_{c,t})\\
    q_{t+1} = q_t + \Delta tv_t\\
    \phi(q_t) \geq 0 ~\bot~ f_n \geq 0\\
    f_n \geq 0  ~\bot~ n^T (V_{t+1} + \epsilon v_t) \geq 0\\
    f_t \geq 0  ~\bot~ \lambda e + D^T v_{t+1} \geq 0\\
    \lambda \geq 0  ~\bot~ \mu f_n - e^T f_t \geq 0
\end{aligned}
\end{equation}

where $M$ is the mass-inertia matrix of the rigid body, $k(v_t, q_t)$ stands for the coriolis and conservative forces terms, $f_{e, t}$ and $f_{c,t}$ are the external and contact forces at time $t$ respectively, $J_c(q_t)$ is the contact Jacobian, $q_{t}$ and $v_t$ are the position and velocity of the rigid body at time $t$, $\phi(q_t)$ is the minimum distance to contact, $f_n$ and $f_t$ are the normal and tangential components of the contact force $f_c$, $\epsilon$ is the coefficient of restitution, $\lambda$ is a scalar that represents the relative tangential velocity term and $\mu$ is the coefficient of friction. 

These equations are rearranged into the LCP form $$V \cdot z + p \geq 0$$  where $V = $ 

\begin{equation}
    \begin{bmatrix}
        \Delta t \cdot n^T M^{-1} n & \Delta t\cdot n^T M^{-1} D & 0\\
        \Delta t \cdot D^T M^{-1} n & \Delta t \cdot D^T M^{-1} D & e \\
        \mu & -e^T & 0
\end{bmatrix}
\end{equation}

$z = $

\begin{equation}
    \begin{bmatrix}
       f_n\\
       f_t\\
       \lambda
    \end{bmatrix}
\end{equation}

and $p = $
\begin{equation}
    \begin{bmatrix}
       n^T ((1+\epsilon)v_t + \Delta t M^{-1} (k +f_e))\\
       D^T (v_t + \Delta t M^{-1}(k + f_e))\\
      0
    \end{bmatrix}
\end{equation}

The solution of this Contact Resolution LCP returns the contact forces, $f_n$ and $f_t$, as well as the relative motion term $\lambda$.

To solve this LCP problem, we redefine it into a Quadratic Program and solve it using Ipopt \cite{ipopt}, which is an off-the-shelf Nonlinear Program Solver that is capable of solving Quadratic Programs. Any off-the-shelf Quadratic Program solver could have been used in place of Ipopt.

The algorithm for contact simulation is summarized in Algorithm \ref{alg:contactsim}. At each time step, a collision checking routine checks for contact. If there is no contact, we simply run a forward simulation of the body's dynamics. If there is contact, the LCP problem is formulated and solved. The solution contact forces are then used to predict the motion of the body in the next time step.

\begin{algorithm}
\caption{Contact Simulation} 
\begin{algorithmic} 
\State Data: Object Geometry, Object Inertial Properties, Friction, Restitution, Initial Conditions
\State Result: Object Trajectory
\State set initial values for $q$ and $v$
\While{$t \leq T$} 
    \State flag = check\_collision($q_t$)
    \If {flag} 
       \State $q_{t+1}, v_{t+1}$ = solve\_LCP($q_t, v_t$)
    \Else 
        \State $v_{t+1} = v_t + \Delta t M^{-1} (k_t + f_{c,t})$
        \State $q_{t+1} = q_t + \Delta t v_t$
    \EndIf
    \State $t = t + 1$ 
\EndWhile
\end{algorithmic}
\label{alg:contactsim}
\end{algorithm}

\subsection{Contact resolution of the circular ball}\label{subsec:ball}
To use the LCP formulation described in Section \ref{subsec:lcp}, to perform contact resolution of the circular ball, all we had to do was to specify our configuration space and specify a function that computed both the contact Jacobian $J_c(q_t)$ and minimum distance to contact $\phi (q_t)$. Since the entire problem is in 2-D space, our body position $q$ and velocity $v$ are in SE(2) and are represented by $(x, y, \theta)$ and $(\dot{x}, \dot{y}, \dot{\theta})$ respectively. Given an initial position and velocity, the trajectory of the ball is simulated using Algorithm \ref{alg:contactsim} above.

\subsection{Contact resolution of 2D Bipedal Robot}\label{subsec:bipedcontact}
We approximate the foot of the 2D bipedal robot as a rectangle of specific width and length to simplify contact resolution. Just like in Section \ref{subsec:ball}, we represent the position and velocity of the foot in SE(2) space and specify a function for computing both the contact Jacobian and minimum distance to contact of the foot. Since the body of the 2D bipedal robot is made up of a torso and legs, both of which have mass, the mass-inertia matrix of the 2D bipedal robot, according to \cite{pranav}, is computed as
$M = $

\begin{equation}
    \begin{bmatrix}
        M_{11} & M_{12} & 0 \\
        M_{21} & M_{22} & 0 \\
        0 & 0 & I \\
    \end{bmatrix}
\end{equation}

where

\begin{equation}
\begin{aligned}
    M_{11} = 2(I  + m_l(c^2 + l^2 - cl - cl cos(\theta))) + m_t l^2\\
    M_{12} = I + m_l(c^2 - cl cos(\theta))\\
    M_{21} = I + m_l(c^2 - cl cos(\theta))\\
    M_{22} = I + c^2 m 
\end{aligned}
\end{equation}

and $m_l$ is the mass of the leg, $c$ is the centroid of the leg, $l$ is the length of the leg, $m_t$ is the mass of the torso and $\theta$ is the angle between the legs.

\subsection{Control of 2D Bipedal Robot}\label{subsec:bipedcontrol}
The 2D bipedal robot has a single actuator situated at the hip that exerts a torque to swing the legs. As a result, the 2D bipedal robot is an underactuated system. We use partial feedback linearization to control the 2D biped robot as described below:

Given the dynamics equation of the 2D bipedal robot as  
\begin{equation}\label{eqn:bipeddynamics}
    M \ddot{\theta} + N = B \cdot u
\end{equation}
where $M$ is the mass-inertia matrix, $N$ is the sum of the coriolis and conservative forces terms, $B$ is the control selection matrix, $\theta = (\theta_1, \theta_2)$ are the 2 degrees of freedom of the robot  and $u$ is the control input, we can rearrange the equation to make $\ddot{\theta}$ the subject of the equation as   
\begin{equation}
    \ddot{\theta} = M^{-1}\cdot (B \cdot u - N)
\end{equation}

Since only the hip joint $\theta_2$ is actuated, we can describe the controlled joint $\theta_c$ as 
\begin{equation}\label{eqn:sc}
\theta_c = S_c \cdot \theta 
\end{equation}

where $S_c$ is the selection matrix $ [0; 1 ]$. Taking the second derivative of Equation \ref{eqn:sc} gives us
\begin{equation}\label{eqn:sel}
    \ddot{\theta_c} = S_c \cdot \ddot{\theta} =  S_c \cdot M^{-1}\cdot (B \cdot u - N)
\end{equation}

Using this trick now makes $\theta_c$ fully controllable. As such we can control $\theta_c$ to any reference value $\theta_c ^{ref}$ using the PD controller 
\begin{equation}\label{eqn:pd}
    \ddot{\theta_c} = \ddot{\theta_c ^{ref}} - K_p (\theta_c - \theta_c ^{ref}) - K_d(\dot{\theta_c} - \dot{\theta_c ^{ref}})
\end{equation}

Putting the expression for $\ddot{\theta_c}$ in Equation \ref{eqn:pd} back into Equation \ref{eqn:sel} and making the control input torque $u$ the subject, we get the expression of the control input torque as 
\begin{equation}\label{eqn:ctrl}
    u = (S_c \cdot M^{-1} \cdot B)^{-1} \cdot (\ddot{\theta_c} + S_c \cdot M^{-1} \cdot N)
\end{equation}

\subsection{Kicking the ball}
To kick the ball with a specified force $F_k$, we find the corresponding torque $\tau_k$ to be applied to the hip actuator of the 2D biped robot using $$\tau_k = F_k \cdot l$$ where $l$ is the length of the leg. $\tau_k$ is then sent as a reference torque to the robot controller, which then computes the control input torque to set the hip actuator as described in Equations \ref{eqn:bipeddynamics} - \ref{eqn:ctrl} above. Once the robot's leg makes contact with the ball, the contact forces are resolved as described above and used to predict the velocity of the ball right after impact. 
\section{Experiments}\label{sec:experiments}
We run the contact simulation of the 2D bipedal robot walking over to the circular ball and kicking it with varying torques. The ball had a mass of $0.04$kg and a diameter of $0.4$m. The robot had a torso mass of $1.0$kg, a leg mass of $0.5$kg and a leg length of $1.0$m. The kicking torques varied from $30$Nm to $100$Nm. The ball trajectories can be seen in the plots in Figure \ref{fig:trajs} below:
\begin{figure}%
    \centering
    \subfloat[\centering ]{\subcaptionbox{30Nm}{\includegraphics[width=.4\linewidth]{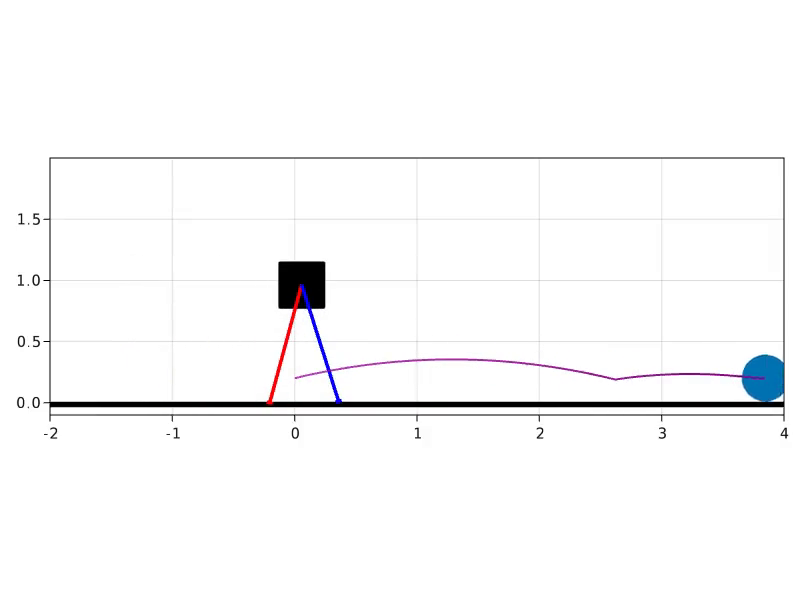} }}%
    \qquad
    \subfloat[\centering ]{\subcaptionbox{40Nm}{\includegraphics[width=.4\linewidth]{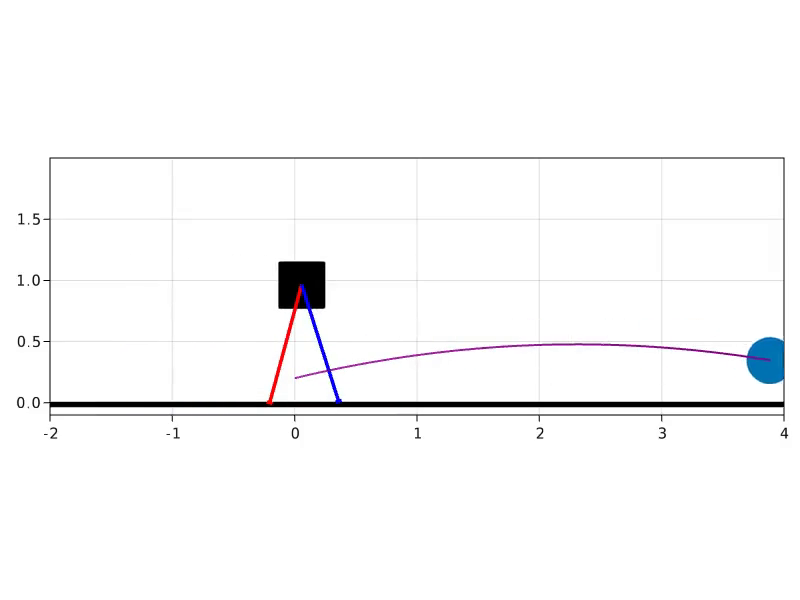} }}%
    \qquad
    \subfloat[\centering ]{\subcaptionbox{50Nm}{\includegraphics[width=.4\linewidth]{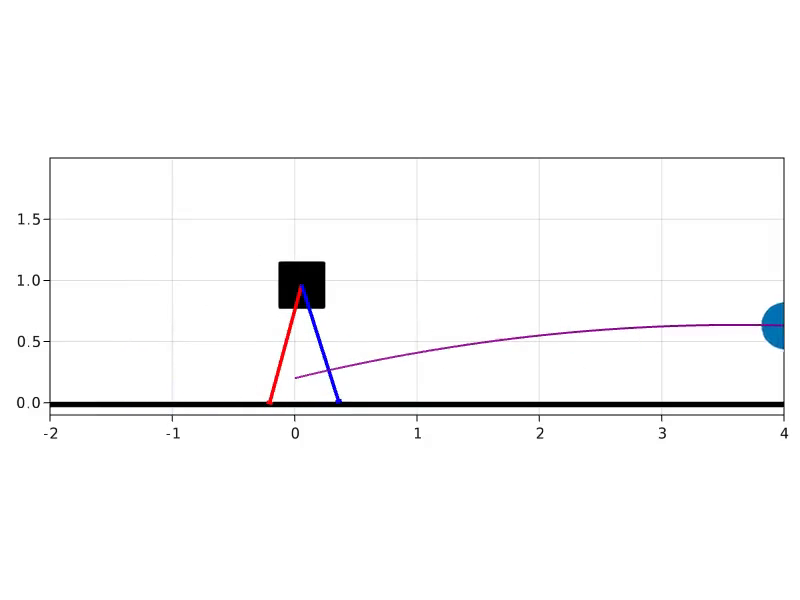} }}%
    \qquad
    \subfloat[\centering ]{\subcaptionbox{60Nm}{\includegraphics[width=.4\linewidth]{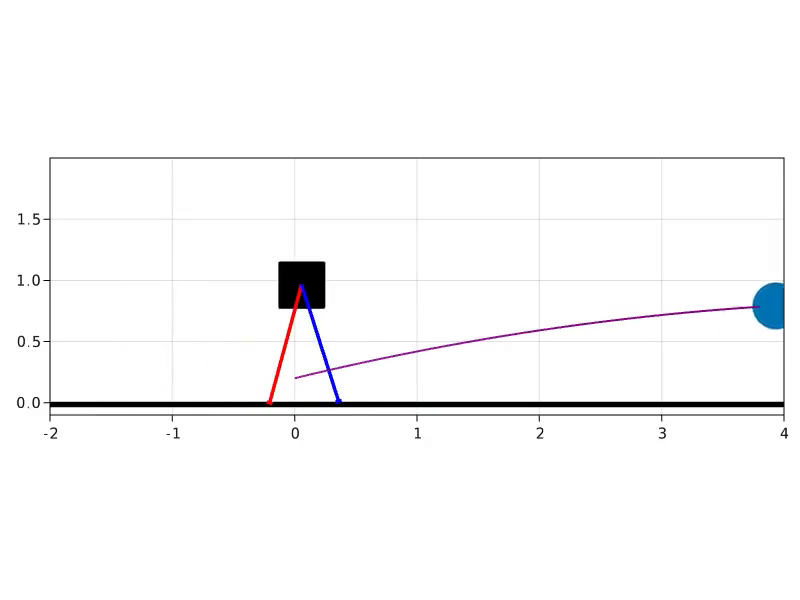} }}%
    \qquad
    \subfloat[\centering ]{\subcaptionbox{70Nm}{\includegraphics[width=.4\linewidth]{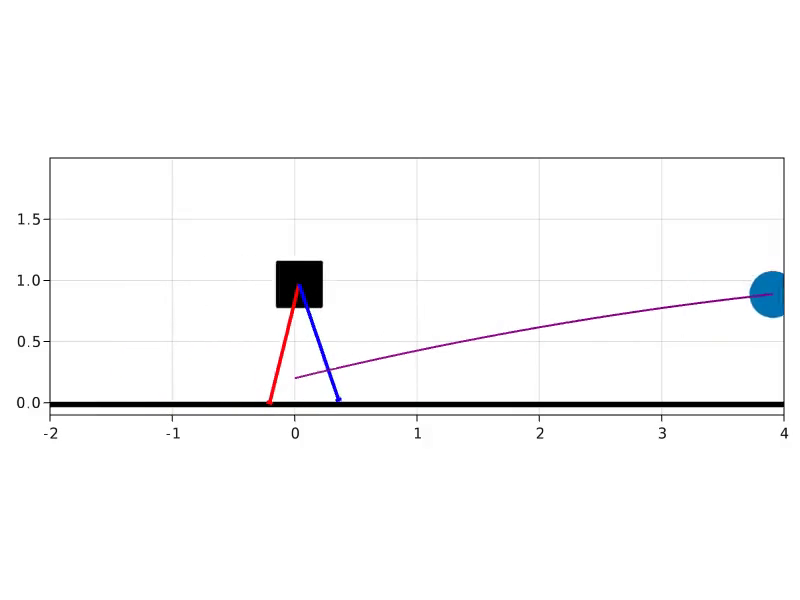} }}%
    \qquad
    \subfloat[\centering ]{\subcaptionbox{80Nm}{\includegraphics[width=.4\linewidth]{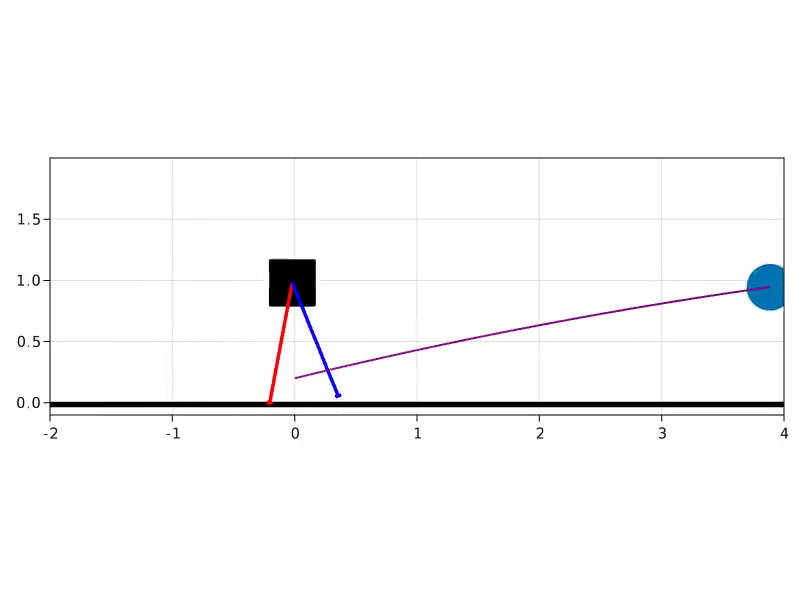} }}%
    \qquad
    \subfloat[\centering ]{\subcaptionbox{90Nm}{\includegraphics[width=.4\linewidth]{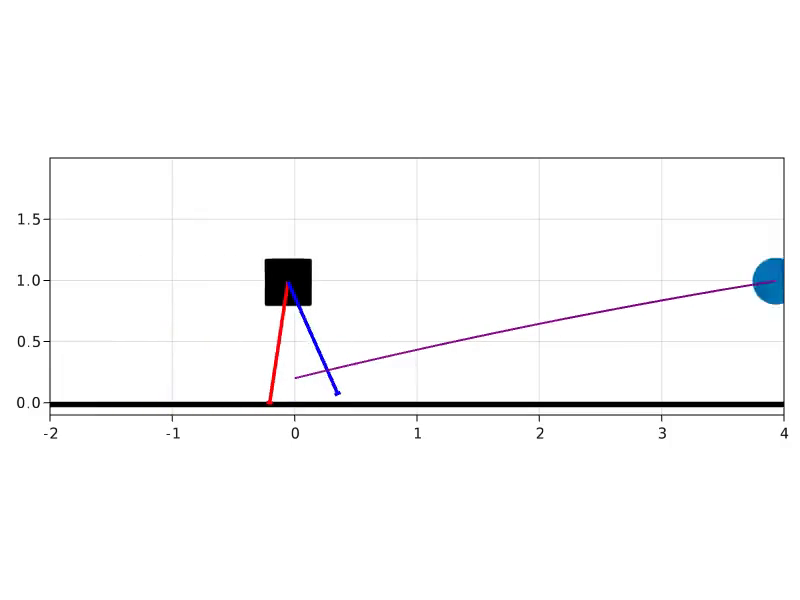} }}%
    \qquad
    \subfloat[\centering ]{\subcaptionbox{100Nm}{\includegraphics[width=.4\linewidth]{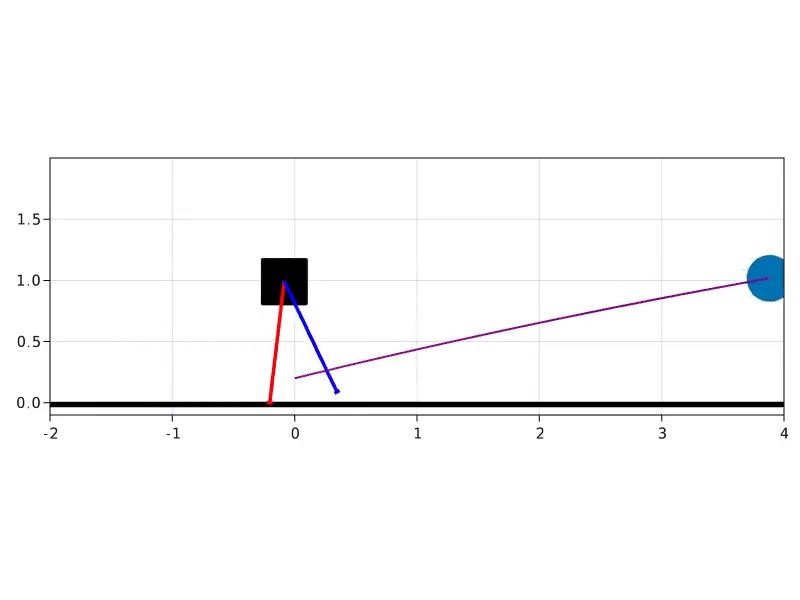} }}%
    \caption{Circular ball trajectories after being kicked by the 2D bipedal robot with torques from 30Nm to 100Nm}%
    \label{fig:trajs}%
\end{figure}

The ball velocities after being kicked with the various leg torques are presented in the table \ref{table:results} below:
\begin{table}
\centering
\begin{tabular}{|c||c|c|} 
\hline  
$\tau$ (N.m)  & $v_x$ (m/s) & $v_y$ (m/s) \\
  \hline\hline
$30.0$ & $7.29$ & $1.78$  \\
\hline
$40.0$ & $9.71$ & $2.38$  \\
\hline
$50.0$ & $12.14$ & $2.97$  \\
\hline
$60.0$ & $14.57$ & $3.57$  \\
\hline
$70.0$ & $17.00$ & $4.16$  \\
\hline
$80.0$ & $19.43$ & $4.75$  \\
\hline
$90.0$ & $21.86$ & $5.35$  \\
\hline
$100.0$ & $24.28$ & $5.94$  \\
\hline
\end{tabular}
\caption{Ball velocities for varying kicking torques after kicking impact  }
\label{table:results}
\end{table} 
\section{Discussion}
The contact simulation of the 2D bipedal robot kicking the circular ball worked quite well. The main difficulty in this work was in implementing the partial feedback linearization controller for the bipedal robot's locomotion. One potential improvement or extension to this work could be to simulate kicking contacts of more irregularly-shaped objects like bottles, buckets or even dogs!. Another extension of this work could be to simulate both arm and leg contacts of a dual-arm bipedal robot pushing a heavy cabinet through a doorway. 
\section{Conclusion}\label{sec:conclusion}
In this report, we described our approach for contact simulation of a 2D bipedal robot kicking a circular ball. We also provided experimental results for the different kicking torques. The code for this work and videos of the contact simulation can be found at this URL: \href{https://github.com/adubredu/contact_sim.jl}{https://github.com/adubredu/contact\_sim.jl}

\bibliographystyle{plain}
\bibliography{references}

\end{document}